\documentclass{article}

\usepackage[preprint]{corl_2026} % Display authors in the arXiv preprint.
\usepackage{multirow}
\usepackage{amsmath,amssymb,amsfonts}
\usepackage{graphicx} 
\usepackage{booktabs}
\usepackage{longtable}
\usepackage{array}
\usepackage{textcomp}
\usepackage{bm}
\usepackage{pifont}
\usepackage{caption}
\usepackage{wrapfig}
\usepackage{enumitem}

\newcommand{\cmark}{\ding{51}}
\newcommand{\xmark}{\ding{55}}
\newcommand{\NA}{--}
\newcommand{\nr}{n.r.}

\newcolumntype{L}[1]{>{\raggedright\arraybackslash}p{#1}}

\title{TAGA: Terrain-aware Active Gaze Learning for Generalizable Agile Humanoid Locomotion}
\author{%
  \makebox[0pt][c]{%
    \begin{minipage}[t]{\textwidth}
      \centering\normalfont
      \setlength{\parskip}{0pt}
      \setlength{\parindent}{0pt}
      \fontsize{9}{11.5}\selectfont
      \textbf{\mbox{LI Peizhuo\textsuperscript{1,*}}, \mbox{Hongyi LI\textsuperscript{2,*}}, \mbox{Mingfeng FAN\textsuperscript{1,*}}, \mbox{Yutong WANG\textsuperscript{1}}, \mbox{Yifeng ZHANG\textsuperscript{1}},\\
        \mbox{Fangzhou XU\textsuperscript{1}}, \mbox{Dapeng JIA\textsuperscript{1}}, \mbox{Shuhao LIAO\textsuperscript{1}}, \mbox{Yuxuan MA\textsuperscript{1}}, \mbox{Zicheng ZENG\textsuperscript{3}}, \mbox{Ze WANG\textsuperscript{2}},\\
        \mbox{Yongbin JIN\textsuperscript{2,\ensuremath{\dagger}}}, \mbox{Yuhong CAO\textsuperscript{1,\ensuremath{\dagger}}}, \mbox{Armin LEDERER\textsuperscript{4}}, \mbox{Hongtao WANG\textsuperscript{2}}, \mbox{Guillaume SARTORETTI\textsuperscript{1}}}\endgraf
      \vspace{2pt}
      \fontsize{9}{10.5}\selectfont
      \textsuperscript{1}MarmotLab, National University of Singapore\quad
      \textsuperscript{2}Center of X-Mechanics, Zhejiang University\\
      \textsuperscript{3}South China University of Technology\\
      \textsuperscript{4}Department of Electrical and Computer Engineering, National University of Singapore\endgraf
      \vspace{1pt}
      \textsuperscript{*}Equal contribution\quad
      \textsuperscript{\ensuremath{\dagger}}Corresponding authors\\
      Project Page: \url{https://marmotlab.github.io/taga-humanoid/}
    \end{minipage}%
  }%
}

\begin{document}
\maketitle

\begin{figure*}[h]
    \centering
     \vspace{-1.0cm}
    \includegraphics[
        width=\textwidth,
    ]{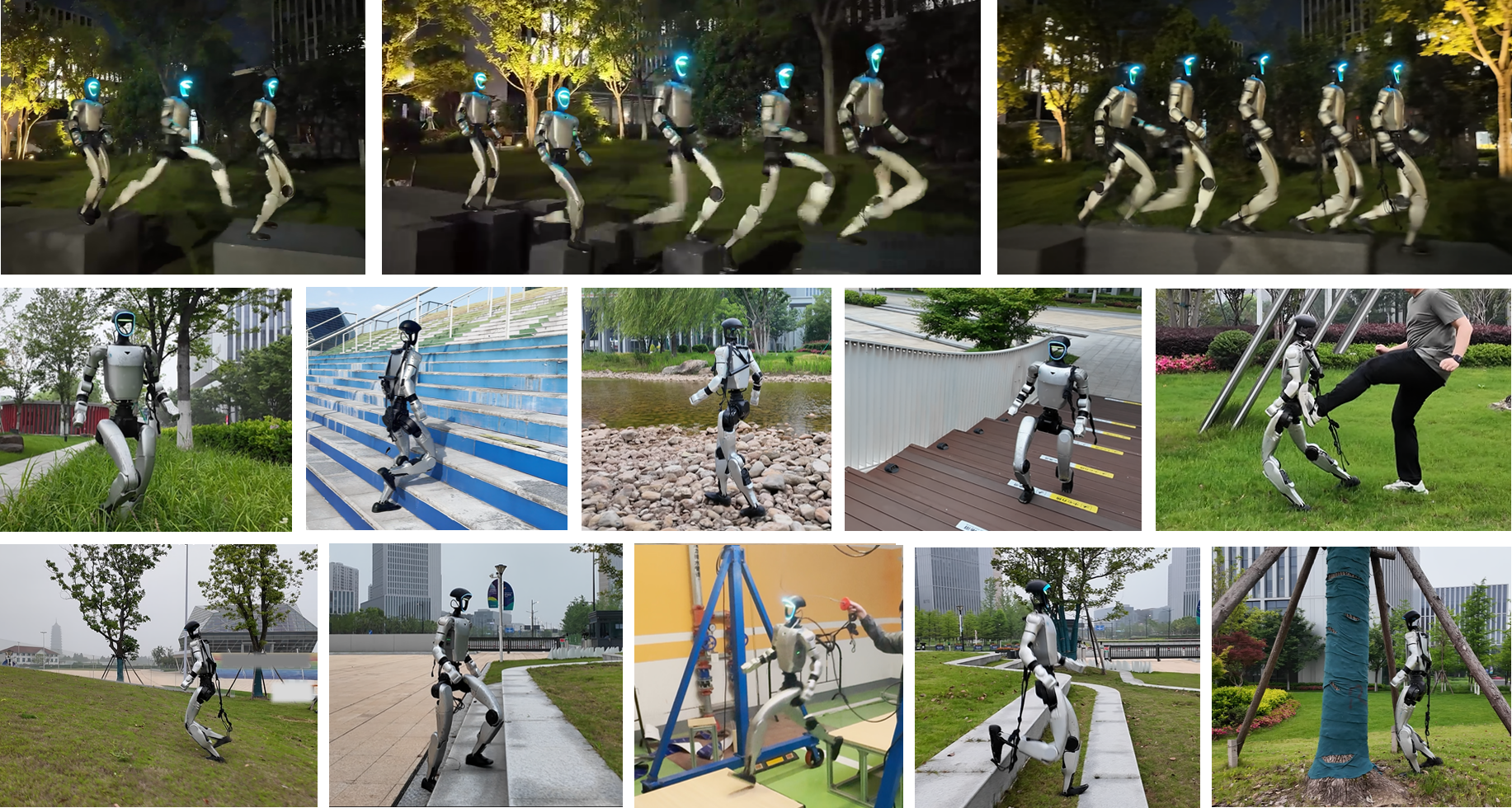}
    \vspace{-0.5cm}
    \caption{
TAGA enables agile and robust humanoid locomotion across diverse challenging terrains.
Deployed on a Unitree G1 with onboard Jetson Orin inference, the robot traverses up to $1.2\,\mathrm{m}$ gaps, narrow beams, sparse stepping stones, stairs, and outdoor terrain.
TAGA uses egocentric vision and proprioception to predict task-relevant regions in the height scan, selectively routing these local terrain observations to the downstream locomotion policy, while remaining robust to severe perceptual disturbances and environmental interference. 
    }
    \vspace{-0.6cm}
    \label{fig:front_page}
\end{figure*}

\begin{abstract}
Agile humanoid locomotion across diverse challenging terrain demands both wide perceptual coverage and precise local geometry understanding. Motivated by the way humans selectively look at relevant terrain during locomotion, we introduce TAGA, a Terrain-aware Active Gaze learning framework for Attention-based humanoid control. By fusing vision, proprioception, and motion commands, our framework guides the model to learn anticipatory cues and actively attend to specific areas of the height scan, selectively using these informative regions for the downstream network. This adaptively increases the information density of observations under tight onboard computational constraints, thus enabling fine-grained perceptive locomotion over larger-scale terrains. We find that such gaze behaviors can naturally emerge through reinforcement learning alone, without requiring additional supervision or explicit guidance, significantly improving training efficiency. As a result, the trained policy demonstrates robust and generalizable locomotion in simulation and on hardware, including reliable terrain-aware foothold selection, elevated-platform traversal, competitive sparse-foothold traversal, and the largest reported real-world gap traversal distance of $1.2\,\mathrm{m}$ among perceptive humanoid locomotion systems, while maintaining stability under severe perceptual disturbances and environmental interference.
\end{abstract}
\keywords{Humanoid Locomotion, Gaze Mechanism, Multimodal Perception} 

%===============================================================================

\section{Introduction}

Humanoids have shown significant advantages over wheeled robots in crossing obstacles, traversing discontinuous terrain, and navigating complex spatial structures. While recent motion-tracking methods enable dynamic whole-body behaviors such as dancing, backflips, and human-motion imitation~\citep{sombolestan2024adaptive, ben2025homie, zhang2025falcon, fey2025bridging, murooka2021humanoid, bouyarmane2018quadratic, fu2024humanplus}, tracking predefined motions is fundamentally different from locomotion in complex terrain, from cluttered indoor scenes to unstructured outdoor landscapes. The former resembles reproducing a memorized trajectory, whereas the latter requires the robot to actively perceive its surroundings, reason about terrain traversability, and adapt its motion strategy in real time~\citep{he2025attention}. This makes perception a central bottleneck: the robot must obtain look-ahead awareness of upcoming terrain while retaining precise local geometry for reliable foothold placement.

\begin{wrapfigure}{r}{0.4\linewidth}
    \centering
    \vspace{-0.43cm}
    \includegraphics[width=\linewidth]{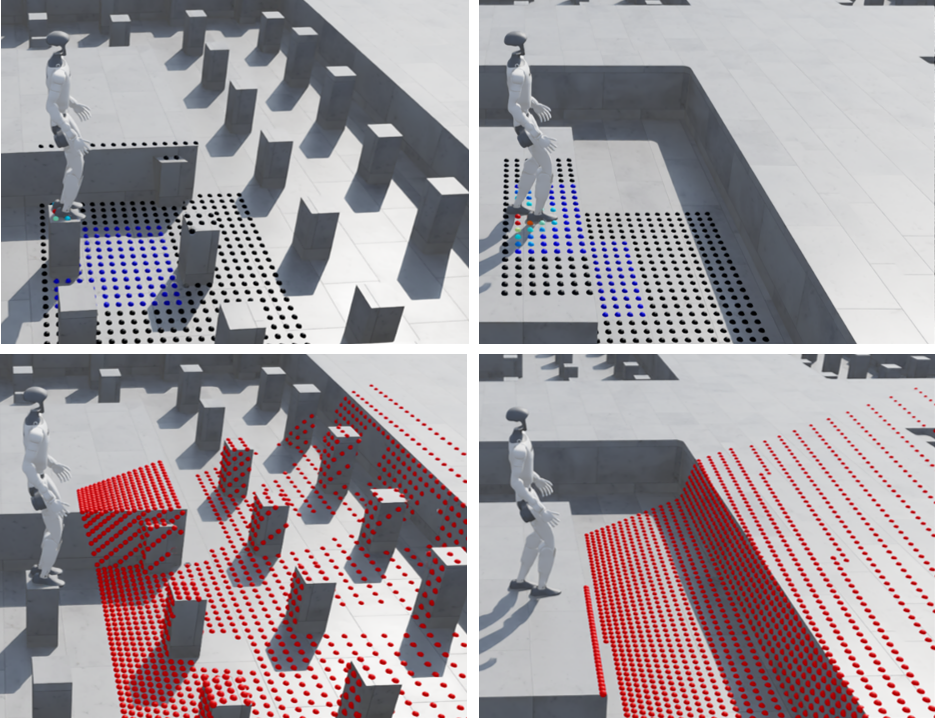}
    \vspace{-10pt}
    \caption{Comparison between local height scan and depth image perception.}
    \label{fig:placeholder}
    \vspace{-10pt}
\end{wrapfigure}

Existing perceptive locomotion methods can generally be divided into two categories: mapping-based methods and vision-based methods. Mapping-based approaches use point clouds or reconstructed height scans as compact terrain representations for locomotion~\citep{he2025attention, miki2022learning, long2025learning, hoeller2024anymal}. While effective, these methods often incur increasing computational cost as the perception range expands. In contrast, vision-based methods directly map raw depth images to actions, reducing reliance on explicit terrain reconstruction~\citep{cheng2024extreme, agarwal2023legged, song2026gait, yang2022learning, zhuang2024humanoid}. However, forward-facing depth images often miss the terrain near or beneath the robot's feet, and recurrent memory struggles to preserve fine-grained geometry over long horizons~\citep{zhang2026ame}.

These two perceptual paradigms actually provide complementary information. As shown in Fig.~\ref{fig:placeholder}, vision offers look-ahead awareness of distant terrain, while a height scan provides accurate local geometry for foot placement and contact-rich motion control. However, many existing approaches either rely on a single source or couple the two only loosely. As a result, they fail to fully exploit the complementary strengths of multimodal perception and cannot effectively align look-ahead visual cues with local geometric details. 
In contrast, humans and animals actively direct their gaze toward task-relevant regions, such as nearby footholds, gaps, or distant obstacles, based on the situation. This suggests that robust humanoid locomotion requires not only multimodal perception, but also an active perception mechanism for deciding \emph{where to look} and \emph{which perceived information matters most for the next step(s)}.

To this end, we propose \textbf{TAGA}, an active perception framework for generalizable agile humanoid locomotion. TAGA is built around a hierarchical gaze mechanism: vision provides long-range terrain preview, height scans supply precise local geometry, and a learned active gaze policy selectively attends to the most locomotion-relevant regions, adaptively fusing these complementary signals. Concretely, TAGA consists of two core components. A \emph{Task-Relevant Active Gaze Module} fuses vision, proprioception, and motion commands to predict which terrain region is most relevant for the next movement and crops the corresponding patch from the height scan. A \emph{Visuomotor Fusion Encoder} then applies cross-attention over this selected region, emphasizing geometric structures critical for foothold placement and terrain-aware decision making. This hierarchical design increases the effective information density of observations, reduces interference from irrelevant terrain, and simplifies downstream policy learning. 
The main contributions of this work are summarized as follows:

\begin{enumerate}[leftmargin=2em,itemsep=0.2em]
    \item We present TAGA, a perceptive locomotion system that integrates depth vision, height scan, and proprioception into a unified sensing and control pipeline for humanoids operating in challenging terrain with onboard computation, while improving training efficiency over full-context.
    \item At the core of TAGA is an emergent hierarchical active gaze mechanism learned without explicit gaze supervision: the policy learns to select a task-relevant terrain patch via visual and proprioceptive cues, then applies fine-grained attention to that region for precise understanding.
    \item We deployed TAGA on Unitree G1 and demonstrated state-of-the-art (SOTA) performance across gaps, stepping stones, narrow beams, stairs, and outdoor terrain. Notably, the robot achieves a 120 cm gap crossing, surpassing the best reported result by 50\%.
\end{enumerate}

\section{Related Works}

\noindent\textbf{Terrain Mapping-Based Perceptive Locomotion.}
A major class of perceptive locomotion methods relies on explicit geometric terrain representations such as height scans, elevation maps, or voxel grids to guide locomotion policies~\cite{fankhauser2018robust,jenelten2020perceptive,wang2025sf,miki2022elevation}. These approaches generally achieve strong local terrain awareness and precise foothold placement in challenging tasks such as stepping stones, sparse footholds, and gap traversal~\cite{dong2025marg,zhang2024learning}. Recent methods further improve terrain understanding through learned geometric encodings, including multi-layer height scans~\cite{chen2025learning}, 3D terrain representations~\cite{miki2024learning}, and Attention-Based Map Encoding (AME)~\cite{he2025attention}. However, mapping-based approaches are usually constrained by the size and resolution of the terrain representation, which limits the tradeoff between local detail and broader terrain coverage~\cite{chen2025learning,miki2024learning}. Their performance also depends on map quality and can degrade under localization uncertainty, occlusions, or deficient perception~\cite{fankhauser2018probabilistic}.

\noindent\textbf{Vision-Based Perceptive Locomotion under Partial Observations.}
Vision-based methods reduce reliance on explicit mapping by learning policies from depth images, RGB-D inputs, or egocentric vision~\cite{agarwal2023legged,yu2021visual,duan2024learning}. 
They have enabled agile locomotion across sparse footholds, discontinuous terrains, complex obstacle traversal, and humanoid stepping tasks~\cite{cheng2024extreme,wang2025beamdojo,rudin2025parkour}, with recent humanoid-oriented works further improving performance through internal models, depth reconstruction, voxel-grid navigation, and limited-view omnidirectional control~\cite{long2025learning,sun2025dpl,ben2025gallant}.
However, these policies often encode terrain reasoning implicitly~\cite{yang2022learning,duan2024learning} and remain vulnerable to partial observability, especially when forward-facing sensors miss underfoot geometry needed for sparse foothold placement~\cite{li2025move,li2025kivi}. Recent methods address this issue using memory or reconstruction, including implicit-explicit learning, visuospatial or volumetric memory, world-model perception, spatial recurrent memory, neural scene representations, sparse-terrain reconstruction, and diffusion-based occupancy synthesis~\cite{luo2024pie,yang2023neural,lai2025world,zhang2024resilient,hoeller2022neural,yu2025start}. Nevertheless, they can still suffer from viewpoint changes, motion uncertainty, and drift between stored terrain representations and the robot's current pose, particularly when latent memory lacks explicit geometric constraints~\cite{yang2025spatially,reed2024scenesense}.

\noindent\textbf{Active Perception for Locomotion.}
Recent studies suggest that robust locomotion depends not only on perceiving the environment, but also on selecting task-relevant terrain regions during traversal. Prior attention-based locomotion methods have shown that selective terrain encoding and exteroceptive--proprioceptive fusion can improve robustness and generalization~\cite{he2025attention}, while Cross-modal Transformers integrate visual and proprioceptive representations for terrain reasoning~\cite{yang2022learning}. More recent adaptive perception methods, such as ADAPT, mainly improve perception robustness by adaptively clipping noisy observations and suppressing perception noise~\cite{shao2026adapt}, while CART selects relevant temporal context for terrain adaptation~\cite{singh2026cart}. Nevertheless, most existing methods still emphasize short-horizon terrain reasoning, while proactive allocation of perception across both nearby footholds and future terrain structures remains underexplored.

\section{TAGA Framework}
\label{sec:mage}

\begin{figure*}[t]
    \centering
     \vspace{-1.0cm}
    \includegraphics[
        width=\textwidth
    ]{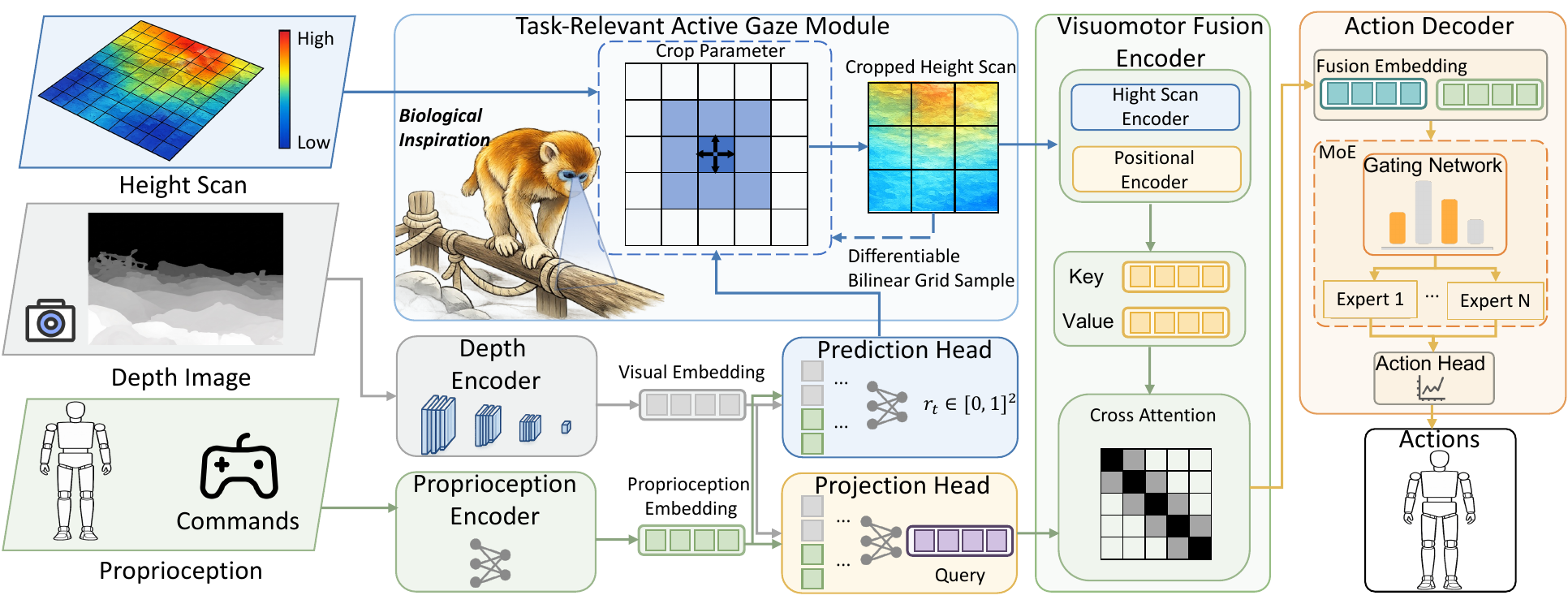}
    \vspace{-0.5cm}
    \caption{The architecture of TAGA.
    }
    \vspace{-0.6cm}
    \label{fig:method}
\end{figure*}

\subsection{Problem Formulation}

We formulate humanoid perceptive locomotion as a partially observable Markov decision process (POMDP; see Appendix~\ref{app:pomdp}). The policy $\pi(a_t|o_t)$ maps the observation $o_t=\{p^H_t,d_t,h_t^{xyz}\}$ to an action $a_t\in\mathbb{R}^{29}$ representing target joint positions, where $d_t\in\mathbb{R}^{1\times36\times64}$ is a forward-facing depth image and $h_t^{xyz}\in\mathbb{R}^{3\times21\times21}$ is a local height scan. The proprioceptive input is a 5-frame history $p^H_t=\{p_{t-4},\ldots,p_t\}$, where each frame is
$p_t=\{\omega_{b,t}, g_{b,t}, q_t, \dot{q}_t, a_{t-1}, c_t\}$.
Here, $\omega_{b,t}\in\mathbb{R}^{3}$ is the measured torso angular velocity, $g_{b,t}\in\mathbb{R}^{3}$ is the projected gravity vector, $q_t,\dot{q}_t\in\mathbb{R}^{29}$ are joint positions and velocities, and $a_{t-1}\in\mathbb{R}^{29}$ is the previous action. The command $c_t=(v^{\mathrm{cmd}}_{x,t},v^{\mathrm{cmd}}_{y,t},\dot{\psi}^{\mathrm{cmd}}_t)\in\mathbb{R}^{3}$ specifies the desired forward velocity, lateral velocity, and yaw rate.

\subsection{Neural Network Design}
\label{sec:taga_nn}

\textbf{Overview.} 
As shown in Fig.~\ref{fig:method}, the TAGA policy $\pi(a_t|o_t)$ is parameterized by a neural network that takes the multimodal observation $o_t$ as input and outputs a vector action $a_t$. TAGA first encodes the depth image $d_t$ and proprioception history $p_t^H$ using a CNN-based depth encoder $\phi_d$ and an MLP-based proprioception encoder $\phi_p$, producing a visual embedding $e^d_t \in\mathbb{R}^{128}$ and a proprioceptive embedding $e^p_t \in \mathbb{R}^{128}$, respectively. The visual embedding captures distant terrain awareness, while the proprioceptive embedding encodes the robot's dynamic state and command context. Given these embeddings and the height scan $h_t^{xyz}$, TAGA employs a hierarchical gaze mechanism to extract locomotion-relevant terrain information. The first stage, the \emph{task-relevant active gaze module}, predicts a region of interest (ROI) in the height scan. The second stage, the \emph{visuomotor fusion encoder}, further emphasizes terrain cues relevant to the next locomotion decision by producing a fusion embedding $e^{pg}_t$ based on the cropped ROI.
Finally, a mixture-of-experts (MoE) based \emph{action decoder} maps the fusion embedding $e^{pg}_t$ and proprioceptive embedding $e^p_t$ to the action output $a_t$. This design enables vision and proprioception to guide where the robot should look, while height scans provide precise local geometry, thereby supporting terrain-aware agile locomotion.

\noindent\textbf{Task-Relevant Active Gaze Module.}
TAGA enables a robot to actively focus its gaze on the ROI via the task-relevant active gaze module. Conditioning on visual preview and proprioceptive state, this module filters irrelevant terrain and directs perception to the most informative region for the next locomotion step.
Concretely, a lightweight prediction head $f_{\mathrm{roi}}$ takes the visual and proprioceptive embeddings as input and predicts a normalized two-dimensional gaze location:~$r_t = f_{\mathrm{roi}}([e^d_t, e^p_t]), r_t \in [0,1]^2$. The predicted location is then mapped onto the full local height-scan grid and used to crop a compact terrain patch: $\tilde{h}^{xyz}_t = \mathrm{Crop}(h^{xyz}_t, r_t), \tilde{h}^{xyz}_t \in \mathbb{R}^{3 \times K \times K}$, where $K=11$ denotes the crop size. The crop is implemented using differentiable bilinear grid sampling, allowing this module to be optimized end-to-end with the policy.
To discourage degenerate solutions near the height-scan boundary, we apply a boundary penalty $\mathcal{L}_{\mathrm{roi}}$ (detailed in Appendix~\ref{app:auxiliary_losses}).

\noindent\textbf{Visuomotor Fusion Encoder.} TAGA further identifies which perceived information is most relevant through the visuomotor fusion encoder. 
Given the cropped patch $\tilde{h}^{xyz}_t$, the encoder first extracts terrain features using a lightweight CNN, while the corresponding spatial coordinates are embedded by an MLP. These features are then fused to obtain pointwise terrain embeddings $E^m_t\in \mathbb{R}^{K \times K \times 128}$, which serve as the key and value vectors. Meanwhile, the query vector is generated from the visual and proprioceptive embeddings through a projection head $f_{proj}$. Finally, the visuomotor fusion encoder applies a multi-head cross-attention layer to obtain the fusion embedding:  $e^{pg}_t = \mathrm{MHA}\bigl(f_{proj}([e^p_t, e^d_t]), E^m_t, E^m_t\bigr)$. The resulting embedding $e^{pg}_t\in \mathbb{R}^{128}$ forms a terrain-aware visuomotor representation for downstream action decoding.

\noindent\textbf{Action Decoder.}
The action decoder is implemented as an MoE module~\citep{ma2026cmoe} to increase policy expressiveness through adaptive action composition. The actor receives the proprioceptive embedding $e^p_t$ and the fusion embedding $e^{pg}_t$ as $e^\pi_t = [e^p_t, e^{pg}_t]$. A gating network $g(\cdot)$ computes soft expert weights $\alpha^i_t = \mathrm{Softmax}(g(e_t^\pi))_i, i=1,\dots,N_e$, and the final action is given by a weighted combination of expert outputs: $a_t = \sum_{i=1}^{N_e} \alpha_t^i \mathcal{E}_i(e_t^\pi)$, where $\mathcal{E}_i$ denotes the $i$-th expert and $N_e=5$ in our implementation. Through soft routing, all experts contribute to action generation with input-dependent weights. This allows the decoder to adaptively compose expert outputs based on TAGA’s task-relevant perceptual and proprioceptive embeddings, improving policy expressiveness for diverse locomotion conditions.

\section{Training TAGA Policy via Reinforcement Learning}
\label{sec:training}
We train TAGA using asymmetric actor-critic PPO in massively parallel simulation~\citep{schwarke2025rslrl}. The actor is described in Sec.\ref{sec:taga_nn}, while the critic additionally receives privileged information, including the ground-truth base linear velocity and the full uncropped height scan.
We support progressive policy learning over risky terrains by constructing multiple terrain types with increasing difficulty and training the policy using curriculum learning. AMP-style regularization and safety-oriented termination further encourage natural gait and stable control. A two-stage training procedure improves both skill acquisition and hardware robustness. Finally, we define the full training objective with symmetry augmentation\citep{mittal2024symmetry} (detailed in Appendix ~\ref{app:auxiliary_losses}).

\noindent\textbf{Terrain Design and Curriculum.}
We design a training terrain set composed of multiple challenging terrain types and train TAGA across these environments. This terrain set covers representative traversal skills, including gap crossing, stair climbing, sparse-foothold traversal, narrow-beam walking, obstacle crossing, and slope locomotion (details in Appendix~\ref{app:terrain_design}), allowing the robot to acquire a broad set of terrain-adaptive locomotion behaviors. Each terrain type is divided into 10 difficulty levels by varying geometric parameters such as gap width, step height, and foothold size. During training, each robot is assigned to a terrain level according to its locomotion performance: successful traversal promotes it to harder terrains, while failure moves it back to easier ones. 

\noindent\textbf{Task Reward and Safety Constraints.}
The base task reward $r^{\mathrm{env}}_t$ combines command tracking, posture and joint regularization, and contact-related terms (detailed in Appendix~\ref{app:reward_terms}). An AMP-style reward $r^{\mathrm{AMP}}_t$ is added to encourage human-like gait, produced by a discriminator trained to distinguish policy motions from motion capture data~\cite{Peng_2021}. We use retargeted AMASS~\cite{AMASS:ICCV:2019} motions as the reference distribution. 
Since extreme agile locomotion may encourage unstable policies, we apply early termination on physically unsafe states, including illegal non-foot-body contacts, excessive torso tilt, insufficient base height, and abnormally large hip-link acceleration during foot contact. These termination conditions serve as hard safety boundaries, filtering out irrecoverable failure cases and focusing learning on stable, feasible locomotion.

\noindent\textbf{Two-stage Training.}
Following AME~\citep{he2025attention}, we use a two-stage procedure to balance skill acquisition and real-world robustness. In the first stage, the policy is trained in a clean setting without observation noise or domain randomization, allowing the robot to acquire core locomotion skills efficiently. Starting from this checkpoint, the second stage fine-tunes the policy with reduced entropy and deployment-oriented randomization, including actuation variations, visual degradation, height-scan noise, external pushes, and terrain perturbations. This improves robustness to noisy real-world sensing and dynamics while preserving the learned active gaze behavior of TAGA.

\noindent\textbf{Loss Function.}
Two auxiliary objectives regularize the learned representations: a contrastive loss $\mathcal{L}_{\mathrm{con}}$ aligning MoE gating latents with full height-scan context to encourage expert specialization~\citep{ma2026cmoe}, and a gaze boundary penalty $\mathcal{L}_{\mathrm{roi}}$ preventing TAGA from degenerately fixating on height-map edges. Putting everything together, the full training objective combines the PPO surrogate loss on the augmented reward $\tilde{r}=r_t^{env}+\eta r_t^{AMP}$ with a value loss $\mathcal{L}_{\mathrm{value}}$, an entropy bonus $\mathcal{H}(\pi_\theta)$, and the auxiliary terms:
\begin{equation}
    \mathcal{L}_{\mathrm{policy}} = \mathcal{L}_{\mathrm{PPO}}(\tilde{r}) + c_v \mathcal{L}_{\mathrm{value}} - c_e \mathcal{H}(\pi_\theta) + \lambda_c \mathcal{L}_{\mathrm{con}} + \lambda_b \mathcal{L}_{\mathrm{roi}},
\end{equation}
where $\eta$ is the AMP reward coefficient, and $c_v$, $c_e$, $\lambda_c$, and $\lambda_b$ are loss weights.

\section{Experiments}
\label{sec:result}
We train TAGA in Isaac Lab~\citep{mittal2025isaaclab} with 8,000 parallel environments on four NVIDIA GeForce RTX 5090 GPUs. The policy is trained over 30k iterations, and then fine-tuned for 10k iterations, requiring about 17 RTX-5090 GPU-days in total. It executes actions at $50\,\mathrm{Hz}$, tracked by a low-level $200\,\mathrm{Hz}$ PD controller. Details are provided in Appendix~\ref{app:hyperparameters}. For real-world experiments, TAGA is deployed on a Unitree G1 humanoid robot with onboard inference on an NVIDIA Jetson Orin. A Livox Mid-360 LiDAR and DLIO provide the local elevation map from which height scans are extracted.

\subsection{Simulation Evaluation}
We first evaluate TAGA in simulation, where terrain conditions and sensing ablations can be controlled systematically. The goal is to characterize its locomotion capability and robustness before real-world deployment. Specifically, we study three questions:
\emph{\textbf{Q1:}} What is the complementary nature of the information provided by visual preview and height scans during challenging terrain traversal?
\emph{\textbf{Q2:}}What are the advantages of our gaze module dynamically selecting task-relevant terrain regions instead of processing the full height scan?
\emph{\textbf{Q3:}} Do AMP priors improve motion naturalness, whole-body coordination, and dynamic stability?
We conduct a comprehensive ablation study for TAGA and compare it with a vision-based baseline. 
\textbf{TAGA-HSOnly} removes depth input and uses only height scans with proprioception. 
\textbf{TAGA-InactiveGaze} deactivates the Gaze Module and uses a fixed ROI. 
\textbf{TAGA-FullScan} removes the Gaze module and processes the full height scan with the same attention-based map encoder. TAGA, TAGA-InactiveGaze, and TAGA-FullScan share the same raw scan; TAGA and TAGA-InactiveGaze use equally sized $11\times11$ ROIs. 
\textbf{TAGA-NoAMP} removes AMP priors.
Since TAGA is designed around local height-scan selection, directly removing height scans would yield an ill-matched vision-only baseline. Instead, we compare TAGA with \textbf{CReF}~\citep{hao2026cref}, a vision-based humanoid locomotion~method.

\begin{table}[t]
\vspace{-1cm}
\centering
\begin{scriptsize}
\setlength{\tabcolsep}{3pt}
\renewcommand{\arraystretch}{1.18}
\captionsetup{width=0.99\textwidth,font=scriptsize}
\caption{
Comparison of TAGA with baseline methods and ablation variants across challenging terrain types. We report GPU count, total training cost in GPU-days, and success rates.
Each method is evaluated over 1000 trials per terrain when applicable.
Bold indicates the best success rate in each terrain column and results within 0.5\% of the best.
}
\label{tab:ablation_success_rate}
\begin{tabular*}{0.99\textwidth}{@{\extracolsep{\fill}}lcccccccc}
\toprule
\textbf{Method} &
\textbf{\# GPUs} $\downarrow$ &
\textbf{GPU-days} $\downarrow$ &
\textbf{Gaps} $\uparrow$ &
\textbf{Stepping Stones} $\uparrow$ &
\textbf{Beam} $\uparrow$ &
\textbf{High Platform} $\uparrow$  &
\textbf{Terrain C1} $\uparrow$ &
\textbf{Terrain C2} $\uparrow$ \\

\midrule
CReF~\citep{hao2026cref}    & 2 & $\sim$10 & 97.40\% & 52.30\% & 96.50\% & 98.70\%  & 85.20\% & 43.10\% \\
TAGA-HSOnly    & 4 & $\sim$17 & 93.10\% & 92.50\% & \textbf{98.30\%} & \textbf{99.60\%} & 90.50\% & 91.50\% \\
TAGA-InactiveGaze & 4 & $\sim$14 & 57.10\% & 83.20\% & 95.60\% & \textbf{100.00\%} & 72.70\% & 48.80\%\\
TAGA-FullScan  & 8 & $\sim$49 & \textbf{99.50\%} & \textbf{98.00\%} & 97.50\% & \textbf{100.00\%} & \textbf{93.40}\% & 92.50\%  \\
TAGA-NoAMP & 4 & $\sim$16 & 96.40\% & 97.20\% & 97.60\% & \textbf{99.50\%} & 89.80\% & 91.60\% \\
\midrule
\textbf{TAGA (Ours)} & 4 & $\sim$17 & 98.30\% & \textbf{97.90\%} & \textbf{98.50\%} & \textbf{100.00\%} & \textbf{93.70\%} & \textbf{93.90\%} \\
\bottomrule
\end{tabular*}
\end{scriptsize}
\vspace{-2mm}
\end{table}

\begin{figure}[t]
    \centering
    \vspace{-2mm}
    \includegraphics[width=1\linewidth]{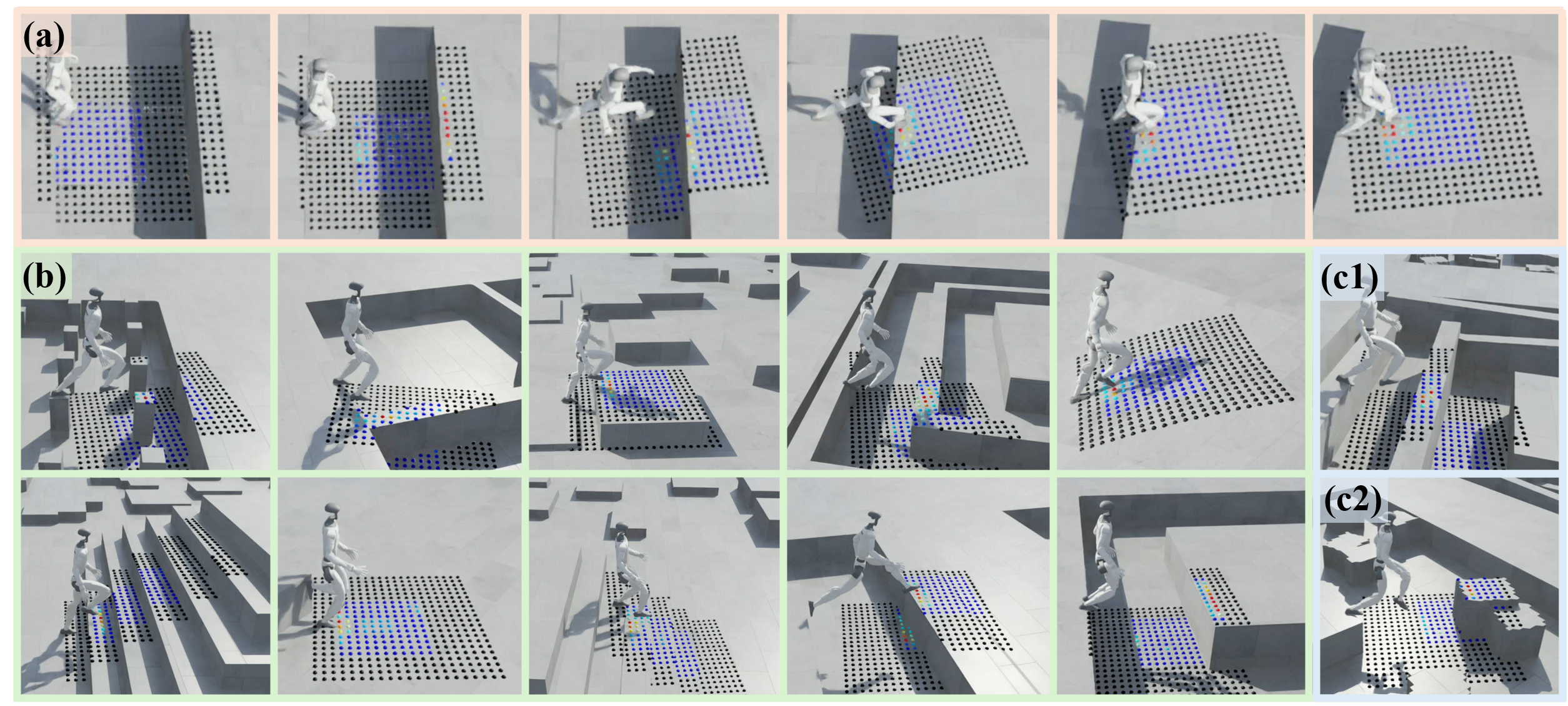}
\caption{
Visualization of the learned active gaze regions and attention-weight distributions of TAGA in simulation.
Black points denote the full local height-scan grid, while colored points indicate the predicted ROI.
Red and blue points indicate higher and lower attention weights, respectively.
(a) shows how the active gaze regions shift during a $1.2\,\mathrm{m}$ gap crossing, and (b) demonstrates attention shifts across diverse terrains. (c) illustrates out-of-distribution testing terrains from training
}
    \label{fig:simulation_gaze}
    \vspace{-3mm}
\end{figure}

\noindent\textbf{Multimodal Perception (Q1).}
Comparing TAGA with CReF and TAGA-HSOnly highlights the complementary roles of visual preview and height scans. TAGA matches or surpasses the vision-based CReF on gaps, beams, and high platforms, and substantially outperforms it on stepping stones (Table~\ref{tab:ablation_success_rate}). This suggests vision alone is insufficient for sparse footholds, where precise foot placement demands accurate local geometry. In contrast, TAGA-HSOnly, which relies solely on height scans and proprioception, struggles with gaps. This indicates that height-scan-only perception lacks sufficient preview to anticipate distant discontinuities and landing regions. Together, these results support the complementary roles of vision and height scans: vision provides anticipatory preview, while height scans provide local geometric precision for robust terrain traversal. Under OOD (out of distribution) terrain challenges (C1, C2), TAGA achieves $93.70\%$ and $93.90\%$, demonstrating stronger robustness and generalization across diverse terrain combinations.

\noindent\textbf{Gaze Module (Q2).}
We assess the effectiveness of our gaze module through visualization (Fig.~\ref{fig:simulation_gaze}) and ablation (Table~\ref{tab:ablation_success_rate}). Instead of using a fixed crop, TAGA predicts task-relevant regions guided by fused visual and proprioceptive cues. During gap crossing, the gaze shifts from the current support region forward to the opposite edge as the robot prepares to cross, resembling the anticipatory gaze behavior observed in humans and animals during locomotion (Fig.~\ref{fig:simulation_gaze}(a)). On stepping stones and beams, it moves toward sparse footholds or the traversable strip; on continuous terrains, the robot's gaze remains local, to cover nearby height changes and contact regions (Fig.~\ref{fig:simulation_gaze} (b)). In our ablations, TAGA achieves performance comparable to TAGA-FullScan with 65.2\% lower training cost than TAGA-FullScan. TAGA-InactiveGaze, with the same compact input but a fixed crop, degrades sharply on gaps and stepping stones where distant footholds lie outside its window.

\begin{wrapfigure}{r}{0.40\linewidth}
    \centering
    \vspace{-3mm}
    \includegraphics[width=1.0\linewidth]{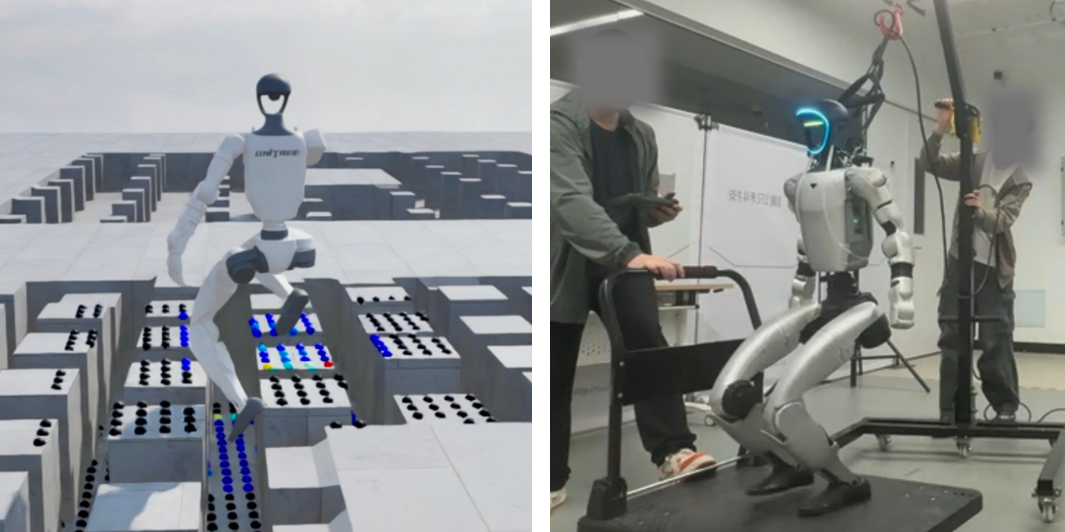}
    \caption{Robot stance without AMP. Without motion guidance, the robot constantly walk with its leg bent.}
    \label{fig:amp}
    \vspace{-0.3cm}
\end{wrapfigure}

\noindent\textbf{Motion Prior (Q3).}
Removing the AMP priors does not substantially hurt task completion, with TAGA-NoAMP performing close to TAGA across all terrains (Table~\ref{tab:ablation_success_rate}). However, Fig.~\ref{fig:amp} shows a clear drop in motion quality. Without AMP, the robot exhibits unnatural motions, including inward knee collapse and short, shuffling steps, especially during turning and terrain transitions. These high-frequency foot motions also made real-world deployment brittle: TAGA-NoAMP proves far less robust to sim-to-real gap and small disturbances.

\subsection{Real-World Evaluation}
\begin{figure}[t]
\vspace{-1mm}
    \centering
    \includegraphics[width=1\linewidth]{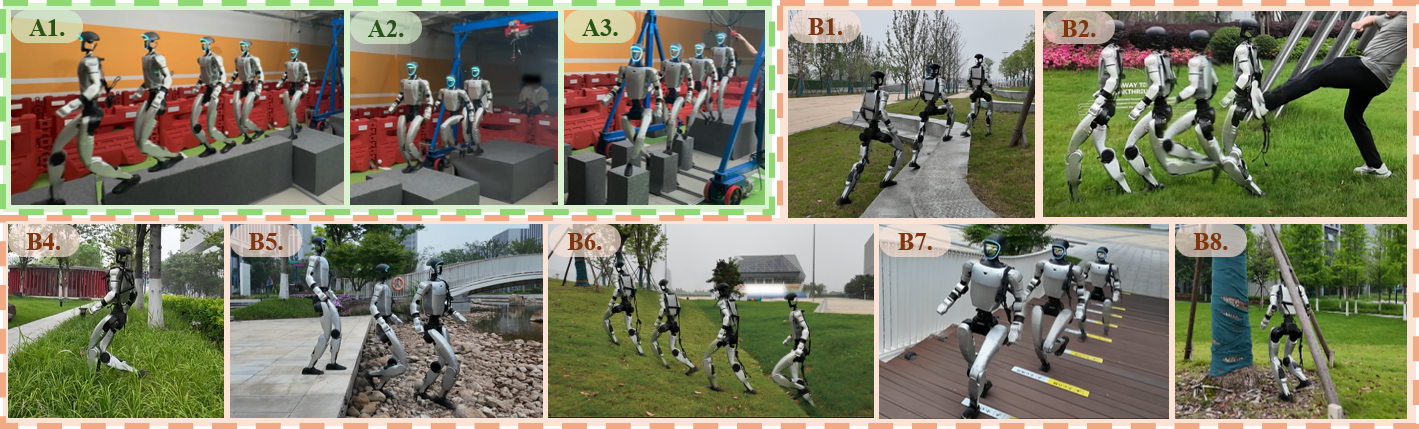}
\caption{
Real-world evaluation of TAGA from controlled indoor terrains to unstructured outdoor environments.
\textbf{(A)} Indoor trials test limit traversal over discontinuous terrains such as platform transitions, large steps, and sparse stepping blocks.
\textbf{(B)} Outdoor trials evaluate robustness to terrain variation, perception degradation, and external disturbances.
Our results show that TAGA maintains stable locomotion across discrete foothold selection, terrain variation, and physical perturbations.}
\label{fig:real_world_exp}
\vspace{-5mm}
\end{figure}

We further evaluate our policy in real-world environments of increasing complexity, moving from controlled indoor trials with designed challenging terrains to outdoor tests involving perception degradation, terrain variation, and physical disturbances. Table~\ref{tab:real_world_perceptive_humanoid_metrics} compares TAGA with reported real-world perceptive humanoid and biped locomotion methods, while Fig.~\ref{fig:front_page} and Fig.~\ref{fig:real_world_exp} show representative indoor and outdoor trials. TAGA covers the widest range of real-world terrain categories among currently reported methods: it \textbf{achieves the largest gap-crossing distance} and can traverse sparse-foothold areas, significantly outperforming previous reported SOTA results on both metrics. Our Vel Tracking AME-2 reproduction adapts the velocity-command interface while retaining the original network structure and terrain-map resolution. We extend the original biped scan's forward coverage to include the opposite gap edge and landing region before takeoff on long gaps, avoiding a comparison limited by unavailable terrain observations.

\begin{table*}[h]
\centering
\begin{scriptsize}
\setlength{\tabcolsep}{2pt}
\renewcommand{\arraystretch}{1.18}
\captionsetup{width=0.92\textwidth,font=scriptsize}
\caption{
Real-world traversal capabilities reported in perceptive humanoid and biped locomotion works.
\nr indicates that the behavior was demonstrated but the terrain geometry was not numerically reported;
\NA indicates that the behavior was not evaluated.
Bold and underline indicate the best and second-best results, respectively.
}
\label{tab:real_world_perceptive_humanoid_metrics}
\begin{tabular}{L{0.2\textwidth} L{0.10\textwidth} L{0.11\textwidth} L{0.12\textwidth} L{0.20\textwidth} L{0.06\textwidth} L{0.06\textwidth}}
\toprule
\textbf{Work} &
\textbf{Robot} &
\textbf{Gap} &
\textbf{Platform} &
\textbf{Sparse foothold} &
\textbf{Beam} &
\textbf{Stairs} \\
\midrule
HPL~\citep{zhuang2024humanoid} &
Unitree H1 &
{\bfseries $\boldsymbol{80\,\mathrm{cm}}$}  &
\underline{$42\,\mathrm{cm}$}&
\NA &
\xmark &
\xmark \\
PIM~\citep{long2025learning} &
Unitree H1 &
\underline{$70\,\mathrm{cm}$}&
{\bfseries $\boldsymbol{50\,\mathrm{cm}}$} &
\NA &
\xmark &
\cmark \\
GA-PHL~\citep{song2026gait} &
LimX Oli &
$46\,\mathrm{cm}$ &
\NA &
\NA &
\xmark &
\cmark \\
AME-1~\citep{he2025attention} &
Fourier GR-1 &
\nr &
\NA &
\nr, uneven &
\cmark &
\cmark \\
\midrule
AME-2~\citep{zhang2026ame} &
LimX Tron1 &
$60\,\mathrm{cm}$  &
{\bfseries $\boldsymbol{48\,\mathrm{cm}}$} &
$40\,\mathrm{cm}$ spacing, uneven &
\cmark &
\cmark \\
Vel Tracking AME-2&
Unitree G1 &
$\underline{90\,\mathrm{cm}}$  &
\underline{$40\,\mathrm{cm}$} &
$40\,\mathrm{cm}$, uneven &
\cmark &
\cmark \\
Now You See That~\citep{sun2026now} &
Honor &
$45\,\mathrm{cm}$ &
\underline{$40\,\mathrm{cm}$}  &
\NA &
\xmark &
\cmark \\
BeamDojo~\citep{wang2025beamdojo} &
Unitree G1 &
$50\,\mathrm{cm}$ &
\NA &
$45\,\mathrm{cm}$ spacing, flat &
\cmark &
\xmark \\
MoRE~\citep{wang2025more} &
Unitree G1 &
$40\,\mathrm{cm}$ &
$30\,\mathrm{cm}$ &
\NA &
\xmark &
\cmark \\
Hiking in the Wild~\citep{zhu2026hiking} &
Unitree G1 &
$50\,\mathrm{cm}$ &
$32\,\mathrm{cm}$ &
\NA &
\xmark &
\xmark \\
Gallant~\citep{ben2025gallant} &
Unitree G1 &
$40\,\mathrm{cm}$ &
$30\,\mathrm{cm}$ &
\NA &
\xmark &
\cmark \\
CReF~\citep{hao2026cref} &
Agibot X2 &
$80\,\mathrm{cm}$  &
\underline{$40\,\mathrm{cm}$} &
\NA &
\xmark &
\cmark \\
RPL~\citep{zhang2026rpl} &
Unitree G1 &
\NA &
\NA &
$60\,\mathrm{cm}$ spacing, flat &
\xmark &
\cmark \\
CMoE~\citep{ma2026cmoe} &
Unitree G1 &
${80\,\mathrm{cm}}$ &
$30\,\mathrm{cm}$ &
\NA &
\xmark &
\cmark \\
\midrule
\textbf{Ours} &
Unitree G1 &
{\bfseries $\boldsymbol{120\,\mathrm{cm}}$} &
\underline{$40\,\mathrm{cm}$} &
{\bfseries $\boldsymbol{70\,\mathrm{cm}}$ spacing, uneven} &
\cmark &
\cmark \\
\bottomrule
\end{tabular}
\end{scriptsize}
\vspace{-2mm}
\end{table*}

\noindent\textbf{Indoor Evaluation.}
Across five hardware trials per terrain, TAGA succeeds in 5/5 trials on stairs, high platforms, narrow beams, and $0.8/1.0\,\mathrm{m}$ gaps, and 4/5 trials on stepping stones and $1.2\,\mathrm{m}$ gaps.
As shown in Fig.~\ref{fig:real_world_exp}\textbf{(A)}, we evaluate TAGA on highly discontinuous indoor terrains with only sparse, separated footholds—long gaps, narrow blocks, and stepping stones—where the robot must carefully plan its footholds and generate sufficient momentum between contacts. TAGA handles these near-limit conditions, crossing $120\,\mathrm{cm}$ gaps, balancing on narrow footholds, and traversing sparse blocks with $50$--$70\,\mathrm{cm}$ spacing and $10\,\mathrm{cm}$ height variation. Beyond these metrics, TAGA exhibits emergent terrain-adaptive behaviors. When crossing long gaps (Fig.~\ref{fig:real_world_exp}\textbf{(A2)}), the robot steps onto the block edge and pushes off for extra momentum rather than taking off from the center of the support region. Upon landing, it bends its knees to absorb impact and quickly recovers balance. These strategies emerge without any hand-crafted motion, indicating that TAGA learns practical locomotion skills for challenging discrete terrains.

\noindent\textbf{Outdoor Evaluation.}
We further evaluate TAGA in outdoor and low-light environments (Fig.~\ref{fig:front_page} and Fig.~\ref{fig:real_world_exp} ), which introduce varied ground materials and challenging perception conditions, including changing illumination, background clutter, depth degradation, sensor noise, as well as occluded footholds in tall grass and confined spaces. Despite these challenges, the robot maintains stable locomotion across terrain types, demonstrating robustness to real-world perception noise and contact variations. Under external disturbances such as kicks and pushes, TAGA absorbs the perturbations, recovers balance, and continues walking without falling. These results indicate that TAGA learns a robust terrain-aware control strategy effective under both perceptual uncertainty and physical disturbances, enabling reliable real-world deployment.
	
%===============================================================================

\section{Conclusion}
\label{sec:conclusion}

We presented TAGA, a terrain-aware active gaze framework for agile humanoid locomotion. TAGA improves traversability across gaps, narrow beams, stairs, elevated platforms, and sparse stepping stones in both simulation and real-world experiments. The learned gaze behavior emerges without explicit supervision, and hardware deployment demonstrates stable locomotion under low-light conditions and strong external disturbances. By fusing vision, proprioception, and motion commands, TAGA selectively attends to task-relevant terrain regions, combining look-ahead visual with local geometric precision for anticipatory, terrain-aware decision-making.

\section{Limitations}

The dynamic maneuvers that TAGA achieves, including explosive jump, precise foothold targeting, and dynamic balance recovery, exert a high thermal load on the actuators during extended operation. This can progressively let actuators overheat, gradually damaging control accuracy and causing reduced jump distance or missed footholds. Additionally, poor height scan quality on complex terrain can cause improper gaits or failures. Future directions include improving tolerance to actuator degradation through online adaptation, better height scan reconstruction, and incorporating uncertainty-aware control to maintain precise foothold selection under degraded hardware and sensing conditions.

\newpage
\bibliography{main}  % .bib

\newpage
\appendix

\section{The Details of POMDP}
\label{app:pomdp}
We formulate humanoid perceptive locomotion as a partially observable Markov decision process (POMDP), denoted as a 6-tuple
$\mathcal{M}=\langle \mathcal{S},\mathcal{O},\mathcal{A},\mathcal{P},\mathcal{R},\gamma\rangle$. Here, $\mathcal{S}$ is the state space, $\mathcal{O}$ is the observation space, $\mathcal{A}$ is the action space, $\mathcal{P}$ is the transition kernel, $\mathcal{R}$ is the reward function, and $\gamma$ is the discount factor.
At each timestep $t$, the robot has an underlying state $s_t\in\mathcal{S}$, which is not fully observable by the policy. The state includes the full robot configuration, velocity, contact state, actuator state, and surrounding terrain geometry. Instead of accessing $s_t$ directly, the policy receives a multimodal observation $o_t\in\mathcal{O}$ composed of proprioception, egocentric depth perception, and local height-scan geometry:
\begin{equation}
    o_t=\{p^H_t,d_t,h^{xyz}_t\}.
\end{equation}
Here, $d_t\in\mathbb{R}^{1\times36\times64}$ denotes the forward-facing depth image, and $h^{xyz}_t\in\mathbb{R}^{3\times21\times21}$ denotes the egocentric height scan, where each grid cell stores an $(x,y,z)$ terrain point. The proprioceptive observation is a five-frame history
\begin{equation}
    p^H_t=\{p_{t-4},\ldots,p_t\},
\end{equation}
where each frame is defined as
\begin{equation}
    p_t=\{\omega_{b,t},g_{b,t},q_t,\dot{q}_t,a_{t-1},c_t\}.
\end{equation}
Specifically, $\omega_{b,t}\in\mathbb{R}^{3}$ is the torso angular velocity, $g_{b,t}\in\mathbb{R}^{3}$ is the projected gravity vector, $q_t,\dot{q}_t\in\mathbb{R}^{29}$ are joint positions and velocities, $a_{t-1}\in\mathbb{R}^{29}$ is the previous action, and $c_t=(v^{\mathrm{cmd}}_{x,t},v^{\mathrm{cmd}}_{y,t},\dot{\psi}^{\mathrm{cmd}}_t)\in\mathbb{R}^{3}$ is the velocity command.

The policy $\pi_\theta(a_t|o_t)$ maps the observation to an action $a_t\in\mathcal{A}$, where $a_t\in\mathbb{R}^{29}$ represents target joint positions at $50\,\mathrm{Hz}$, tracked by a low-level PD controller at $200\,\mathrm{Hz}$. After the action is executed, the environment transitions via $\mathcal{P}(s_{t+1}|s_t,a_t)$ and yields a reward $\tilde{r}_t=\mathcal{R}(s_t,a_t)$ that encourages velocity tracking, stable posture, safe contacts, and human-like motion (see Appendix~\ref{app:reward_terms} for the full reward specification). The complete observation and action spaces are detailed in Table~\ref{tab:obs_act_specs}. The objective is to maximize the expected discounted return:
\begin{equation}
    \max_{\theta}\; \mathbb{E}_{\pi_\theta}
    \left[
    \sum_{t=0}^{T-1}\gamma^t \tilde{r}_t
    \right].
\end{equation}

\begin{table}[h]
\centering
\caption{Action and observation space specifications.}
\label{tab:obs_act_specs}
\begin{tabular}{@{}lll@{}}
\toprule
\textbf{Modality} & \textbf{Symbol} & \textbf{Description} \\
\midrule
\multicolumn{3}{@{}l}{\textit{Action}} \\
\quad Joint position targets & $a_t \in \mathbb{R}^{29}$ & 50 Hz control output, tracked via PD at 200 Hz \\
\midrule
\multicolumn{3}{@{}l}{\textit{Proprioception}} \\
\quad Base angular velocity & $\omega_{b,t} \in \mathbb{R}^{3}$ & Torso angular velocity \\
\quad Projected gravity & $g_{b,t} \in \mathbb{R}^{3}$ & Gravity direction in torso frame \\
\quad Joint positions & $q_t \in \mathbb{R}^{29}$ & Measured joint angles \\
\quad Joint velocities & $\dot{q}_t \in \mathbb{R}^{29}$ & Measured joint velocities \\
\quad Previous action & $a_{t-1} \in \mathbb{R}^{29}$ & Last commanded joint targets \\
\quad Velocity command & $c_t \in \mathbb{R}^{3}$ & $(v^{\mathrm{cmd}}_x, v^{\mathrm{cmd}}_y, \dot{\psi}^{\mathrm{cmd}})$ target \\
\midrule
\multicolumn{3}{@{}l}{\textit{Depth}} \\
\quad Depth image & $d_t \in \mathbb{R}^{1 \times 36 \times 64}$ & Forward-facing, long-range terrain preview \\
\midrule
\multicolumn{3}{@{}l}{\textit{Height Scan}} \\
\quad Height map & $h^{xyz}_t \in \mathbb{R}^{3 \times 21 \times 21}$ & Egocentric $(x,y,z)$ terrain grid, local surroundings \\
\bottomrule
\end{tabular}
\end{table}

\section{Definition of Gaze Learning}
\label{app:gaze}
We define \textit{gaze} as a combination of a normalized gaze location $r_t=(r_t^x, r_t^y)\in[0,1]^2$ and an ROI centered at the gaze location. 
The ROI is defined as a terrain patch 
$\tilde{h}_t^{xyz}\in\mathbb{R}^{3\times K\times K}$ over the local height scan $h_t^{xyz}\in\mathbb{R}^{3\times M\times M}$ with $M=21$. We crop the ROI based on the gaze location, where the normalized gaze coordinates are first mapped to the height scan grid as $(u_t,v_t)=\left(\lfloor r_t^x M\rfloor,\lfloor r_t^y M\rfloor\right)$, and the ROI is obtained by extracting terrain points through
$
\tilde{h}_t^{xyz}
=
\left\{
h_t^{xyz}(i,j)
\;\middle|\;
|i-u_t|\le\frac{K}{2},
\;
|j-v_t|\le\frac{K}{2},
\;
i,j\in\mathbb{Z}
\right\}.
$
Consequently, we define \textit{gaze learning} as training the gaze prediction head $f_{\mathrm{roi}}(\cdot)$ to output the gaze location $r_t=f_{\mathrm{roi}}([e_t^d,e_t^p])$, which is then used to extract an ROI $\tilde{h}_t^{xyz}=\mathrm{Crop}(h_t^{xyz},r_t)$, where $\mathrm{Crop}(\cdot)$ denotes the ROI cropping operation defined above. The extracted ROI $\tilde{h}_t^{xyz}$ is subsequently provided to the downstream locomotion policy for decision-making.

\section{Training Details}
\label{app:training_detail}

\subsection{Terrain Design}
\label{app:terrain_design}
\begin{figure}[h]
    \centering
    \includegraphics[width=1\linewidth]{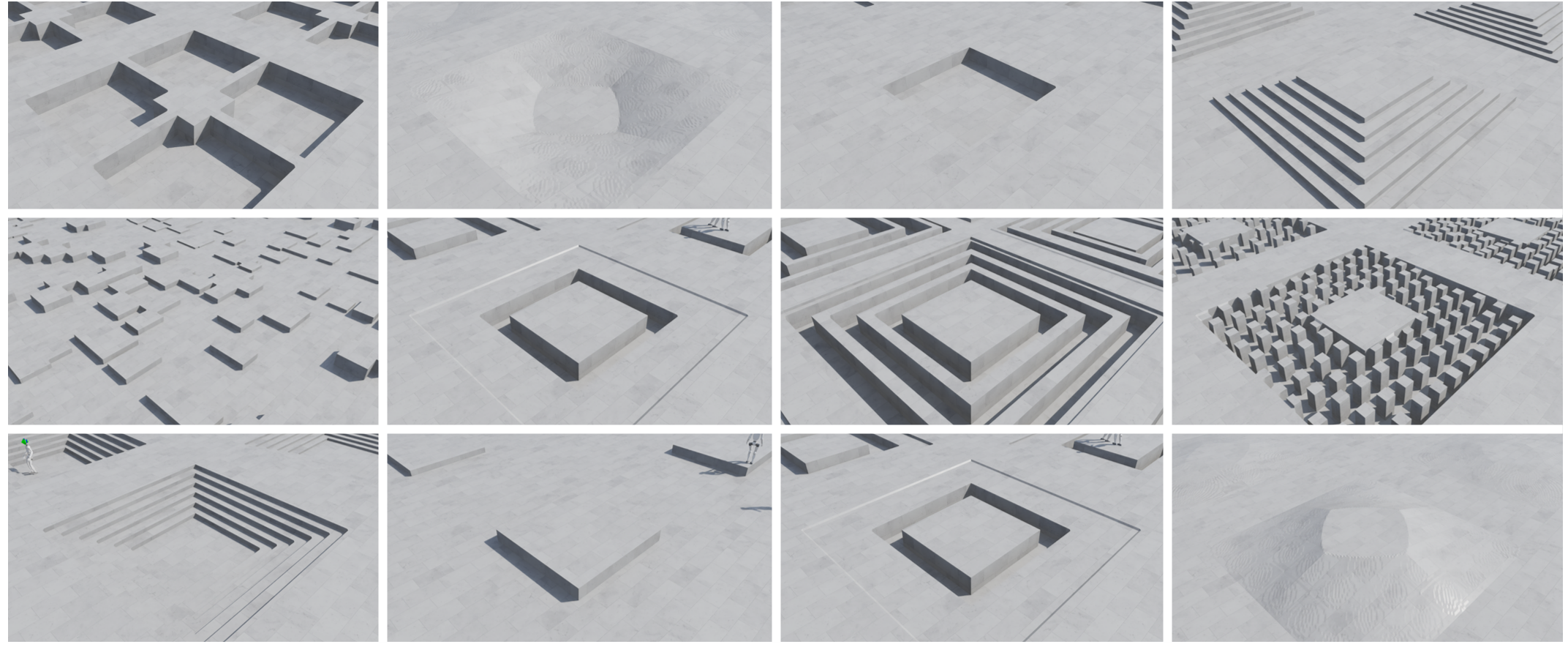}
    \caption{Training Terrains for TAGA}
    \label{fig:terrains}
\end{figure}

To improve traversability across diverse terrain conditions, we construct a broad set of representative terrain types during training, including ascending and descending stairs, gaps, stepping stones (sparse footholds), box obstacles, elevated platforms, and sloped surfaces, as illustrated in Fig.~\ref{fig:terrains}. This terrain mixture exposes the policy to  diverse geometric features, including elevation changes, discontinuous support regions, sparse footholds, obstacle negotiation, and inclined surfaces, thereby encouraging the emergence of generalizable terrain-adaptive locomotion strategies.

Terrain sampling during training follows a probabilistic strategy to balance exposure across terrain types. Stepping stones are sampled with a probability of 0.3, as they most directly exercise precise foothold selection and sparse-terrain reasoning. Ascending stairs, descending stairs, upslope terrains, and downslope terrains are each sampled with a probability of 0.05, since they mainly introduce structured elevation variations and are less demanding in terms of sparse foothold selection. The remaining probability mass is evenly distributed among gaps, box obstacles, and elevated platforms. This sampling strategy allows the policy to experience a wide range of terrain geometries and improves its adaptability to challenging deployment scenarios.

\subsection{Reward Terms}
\label{app:reward_terms}

The locomotion policy is trained with an augmented reward $\tilde{r}_t = r_t^{\mathrm{env}} + \eta r_t^{\mathrm{AMP}}$, where $r_t^{\mathrm{env}}$ denotes the base task reward, $r_t^{\mathrm{AMP}}$ denotes the Adversarial Motion Prior (AMP) reward, and $\eta$ controls the contribution of the AMP term. The base task reward combines command-tracking objectives, posture regularization, joint-level regularization, contact-aware locomotion terms, and safety-oriented penalties. It is designed to encourage agile yet stable locomotion with coordinated contacts and physically plausible motion patterns.

The primary task objective is velocity tracking, where the robot is rewarded for matching the commanded linear and angular velocities. To improve stability and motion quality, we additionally penalize excessive torso rotation, body tilt, joint torques, joint velocities, joint accelerations, action-rate changes, and deviations from nominal body configurations. These regularization terms encourage smooth and energy-efficient motions while reducing unnecessary body oscillations.

To promote robust locomotion behaviors, several contact-related rewards and penalties are introduced. These include rewards for maintaining appropriate foot swing durations and penalties for foot sliding, stumbling, undesired body contacts, loss of ground contact, and excessively narrow foot placement. Together, these terms encourage coordinated stepping behaviors and stable support transitions on challenging terrain. During the second-stage fine-tuning process, additional penalties are activated to improve robustness under deployment-oriented disturbances and terrain variations. 

The detailed components and coefficients of $r_t^{\mathrm{env}}$ are summarized in Table~\ref{tab:reward_terms}. Here, $c^{xy}_t=(v^{\mathrm{cmd}}_{x,t},v^{\mathrm{cmd}}_{y,t})$ and $c^\omega_t=\dot{\psi}^{\mathrm{cmd}}_t$ denote the commanded linear and angular velocity components; $\mathcal{F}$ denotes the set of feet, $\mathcal{B}$ denotes the set of robot links, $q^0$ is the default joint configuration, $f^b_t$ is the contact force acting on body $b$, and $[\cdot]_+=\max(\cdot,0)$ denotes the positive-part operator. The gating function $G_{\mathrm{flat}}(h_t)$ activates the torso-orientation penalty only when the local terrain is sufficiently flat, preventing unnecessary penalization on uneven terrain that naturally requires body inclination.

In addition to these handcrafted reward terms, an AMP reward is used throughout training to encourage human-like locomotion styles. The discriminator is trained using motion-capture data and provides an auxiliary reward that regularizes the learned behaviors toward natural gait patterns.
Specifically, we use retargeted AMASS~\citep{AMASS:ICCV:2019} motions as the reference motion distribution. Both the simulated robot motion and the reference motion are encoded into a compact motion descriptor containing body motion, body orientation, and joint-state information. The discriminator is trained to distinguish policy-generated motion descriptors from AMASS reference descriptors, while the policy is rewarded when its motion is classified as closer to the reference distribution. Unlike frame-wise imitation, this adversarial formulation provides a distribution-level style prior, allowing the policy to preserve task performance while producing smoother and more natural locomotion patterns.

\begin{table*}[t]
\centering
\caption{Reward terms used for locomotion training. The formulas denote the unweighted reward terms. Values in parentheses indicate coefficients used during the second-stage fine-tuning.}
\label{tab:reward_terms}
\small
\resizebox{\textwidth}{!}{
\begin{tabular}{l l l c}
\toprule
Category & Term & Formula & Weight \\
\midrule

Task Objective
& Alive
& $\mathbb{I}_{\mathrm{alive}}$
& $3.0$ \\

& Linear velocity tracking
& $\exp\!\left(-\|v^{xy}_{\mathrm{torso},t}-c^{xy}_t\|^2/\sigma_v^2\right)$
& $2.0$ ($2.5$) \\

& Yaw velocity tracking
& $\exp\!\left(-(\omega^z_{\mathrm{torso},t}-c^\omega_t)^2/\sigma_\omega^2\right)$
& $3.0$ \\

& Yaw command regularization
& $|c^\omega_t|$
& $-1.0$ \\

& Forward progress
& $\mathbb{I}(c^x_t>0.3)\bigl[\mathbb{I}(v^x_{\mathrm{torso},t}<0.15)+\mathbb{I}(v^x_{\mathrm{torso},t}<0)+\mathbb{I}(v^x_{\mathrm{torso},t}<-0.15)\bigr]$
& $-0.5$ \\

\midrule

Posture Regularization
& Torso angular velocity
& $\|\omega^{xy}_{\mathrm{torso},t}\|^2$
& $-0.05$ \\

& Torso orientation
& $G_{\mathrm{flat}}(h_t)\,\|g^{xy}_{\mathrm{torso},t}\|^2$
& $-2.0$ \\

& Pelvis orientation
& $\|g^{xy}_{\mathrm{pelvis},t}\|^2$
& $-0.5$ \\

\midrule

Joint Regularization
& Joint torque
& $\|\tau_t\|^2$
& $-1.5{\times}10^{-7}$ \\

& Joint velocity
& $\|\dot{q}_t\|^2$
& $-5.0{\times}10^{-4}$ \\

& Joint acceleration
& $\|\ddot{q}_t\|^2$
& $-1.25{\times}10^{-7}$ \\

& Link acceleration
& $\frac{1}{|\mathcal{B}|}\sum_{b\in\mathcal{B}}\|\dot{v}^b_t\|$
& $-0.01$ \\

& Hip deviation
& $\|q^{\mathrm{hip}}_t-q^{\mathrm{hip},0}\|^2$
& $-0.1$ \\

& Arm deviation
& $\|q^{\mathrm{arm}}_t-q^{\mathrm{arm},0}\|_1$
& $-0.3$ \\

& Waist deviation
& $\|q^{\mathrm{waist}}_t-q^{\mathrm{waist},0}\|_1$
& $-1.0$ \\

& Joint position limit
& $\sum_j \mathrm{dist}(q^j_t,\mathcal{Q}^j)$
& $-5.0$ \\

& Joint velocity limit
& $\sum_j \left[|\dot{q}^j_t|-\rho_{\dot q}\dot q^j_{\max}\right]_+$
& $-1.0$ \\

& Joint torque limit
& $\sum_j \left[|\tau^j_t|-\rho_\tau\tau^j_{\max}\right]^2_+$
& $-0.01$ \\

& Action rate
& $\|a_t-a_{t-1}\|^2$
& $-0.005$ \\

\midrule

Contact and Gait
& Undesired contact
& $\sum_{b\notin\mathcal{F}}\mathbb{I}(\|f^b_t\|>\epsilon_f)$
& $-1.0$ \\

& Foot air time
& $\mathbb{I}_{\mathrm{cmd}}\mathbb{I}_{\mathrm{single}}\min_{f\in\mathcal{F}}T^f_{\mathrm{mode}}$
& $0.25$ \\

& Air/contact time variance
& $\mathrm{Var}_{f\in\mathcal{F}}(\mathrm{clip}(T^f_{\mathrm{air}},0.5))+\mathrm{Var}_{f\in\mathcal{F}}(\mathrm{clip}(T^f_{\mathrm{contact}},0.5))$
& $-0.7$ ($-2.0$) \\

& Foot stumble
& $\mathbb{I}\!\left(\exists f\in\mathcal{F}: \|f^{xy}_t\|>4|f^z_t|\right)$
& $-2.0$ ($-5.0$) \\

& Foot slide
& $\sum_{f\in\mathcal{F}}\mathbb{I}^f_{\mathrm{contact}}\|v^{xy,f}_t\|$
& $-0.1$ \\

& Foot orientation
& $\sum_{f\in\mathcal{F}}\mathbb{I}^f_{\mathrm{contact}}\|g^{xy,f}_t\|$
& $-0.5$ \\

& No-fly
& $\mathbb{I}\!\left(\sum_{f\in\mathcal{F}}\mathbb{I}^f_{\mathrm{contact}}=0\right)$
& $-2.0$ \\

& Feet too near
& $\left[d_{\min}-\|x_t^L-x_t^R\|\right]_+$
& $-1.0$ ($-5.0$) \\

\midrule

Fine-tuning Only
& Volume penetration
& $\sum_{x\in\mathcal{V}}\mathbb{I}(\delta_x>0)(\|v_x\|+\epsilon)\delta_x$
& $0.0(-1.0)$ \\

& Stand still
& $\mathbb{I}(\|c_t^{xy}\|<\epsilon_c)\mathbb{I}(|c_t^\omega|<\epsilon_c)(\|q_t-q^0\|_1-b)$
& $0.0(-0.3)$ \\

\bottomrule
\end{tabular}}
\end{table*}

\subsection{Training Hyperparameters}
\label{app:hyperparameters}

Table~\ref{tab:training_hyperparameters} summarizes the main PPO optimization hyperparameters used for training TAGA. Unless otherwise specified, the same hyperparameters are used throughout all experiments. Values shown in parentheses correspond to the second-stage fine-tuning configuration when different from the first-stage training setup.

\begin{table}[t]
\centering
\caption{Main training hyperparameters. Values in parentheses indicate the second-stage fine-tuning configuration when different from the first stage.}
\label{tab:training_hyperparameters}
\small
\begin{tabular}{lc}
\toprule
\textbf{Parameter} & \textbf{Value} \\
\midrule
Rollout length & 24 \\
PPO epochs & 5 \\
Mini-batches & 10 \\
Learning rate & $1\times10^{-3}$ \\
Learning-rate schedule & Adaptive \\
Discount factor $\gamma$ & 0.99 \\
GAE parameter $\lambda$ & 0.95 \\
PPO clipping coefficient & 0.2 \\
Value loss coefficient & 1.0 \\
Entropy coefficient & 0.005 (0.002) \\
Gradient clipping norm & 1.0 \\
\bottomrule
\end{tabular}
\end{table}

\subsection{Termination Conditions}
\label{app:termination}

To facilitate stable training and avoid unsafe failure modes, we use the following episode termination conditions during training:

\begin{enumerate}[leftmargin=2em,itemsep=0.2em]
    \item \textbf{Timeout and terrain boundary.}
    An episode is reset after the maximum episode length of $20\,\mathrm{s}$, or when the robot moves outside the valid terrain region.

    \item \textbf{Illegal non-foot contact.}
    We terminate the episode when non-foot bodies make contact with the environment. The monitored links include the torso, pelvis, waist, shoulder links, elbow links, hip links, and knee links. Contacts are detected using a force threshold of $1\,\mathrm{N}$.

    \item \textbf{Bad torso orientation.}
    The episode is terminated when the torso orientation deviates excessively from upright. In implementation, this is measured by the angle between the torso-frame projected gravity direction and the upright axis, with a threshold of $0.8\,\mathrm{rad}$.

    \item \textbf{Low base height.}
    We terminate states in which the robot has effectively fallen or collapsed. This includes cases where the root height is below $0.5\,\mathrm{m}$ relative to the terrain origin, or the root clearance above the local terrain estimate is below $0.2\,\mathrm{m}$.

    \item \textbf{High hip-link acceleration during foot contact.}
    To filter out high-impact failure cases, we terminate the episode when a foot is in contact and the corresponding hip-pitch link experiences excessive linear acceleration. A contact is considered active when the foot contact force exceeds $1\,\mathrm{N}$, and termination is triggered when the hip-link acceleration exceeds $225\,\mathrm{m/s^2}$. This condition discourages stiff, high-impact landings and encourages recoverable contact behaviors.
\end{enumerate}

Triggering any unsafe termination condition resets the episode immediately. These conditions prevent the policy from exploiting unstable behaviors and focus learning on recoverable locomotion strategies.

\subsection{Two-stage Training and Domain Randomization}
\label{app:two_stage_domain_randomization}

We use a two-stage training procedure to balance skill acquisition and real-world robustness. In the first stage, TAGA is trained in a clean simulation setting without observation noise or deployment-oriented dynamics randomization, allowing the policy to efficiently acquire the core locomotion skills. In the second stage, we fine-tune from the pre-trained checkpoint with a reduced entropy coefficient and deployment-oriented domain randomization.

For robot dynamics, we randomize the added base payload within $[-1.0, 3.0]\,\mathrm{kg}$, perturb the base center of mass by $[-0.05, 0.05]\,\mathrm{m}$ along the $x$ and $y$ axes and along the $z$ axis, and sample motor delays from $0$--$3$ delay steps. Contact properties are randomized with static friction in $[0.3, 1.0]$, dynamic friction in $[0.3, 0.8]$, and restitution in $[0.0, 0.5]$. We also apply random pushes every $10$--$15\,\mathrm{s}$ with planar velocity perturbations in $[-0.5, 0.5]\,\mathrm{m/s}$.

For observations, we add independently sampled uniform noise to proprioceptive channels, with maximum magnitudes of $0.2\,\mathrm{rad/s}$ for base angular velocity, $0.05$ for projected gravity, $0.01\,\mathrm{rad}$ for joint positions, and $1.5\,\mathrm{rad/s}$ for joint velocities. The depth camera is updated at $30\,\mathrm{Hz}$ during fine-tuning and is degraded with contour corruption, random depth artifacts, reflections, sky artifacts, Gaussian blur, stereo failure for too-close surfaces, and robot self-occlusion. Depth values are clipped to $[0.4, 3.0]\,\mathrm{m}$ and then normalized before being fed to the policy.

For height scans, we add $z$-axis noise in $[-0.05, 0.05]\,\mathrm{m}$ and ray-cast drift in $[-0.05, 0.05]\,\mathrm{m}$ along each axis. We also randomly corrupt a small portion of height-scan returns to simulate missing or unreliable terrain measurements. Terrain geometry is further perturbed with Perlin roughness, randomized gap-edge transition widths, and virtual edge obstacles. These randomizations improve robustness to perception noise, actuation uncertainty, contact variation, and terrain irregularities that may arise during hardware deployment.

\section{Training Objectives}
\label{app:auxiliary_losses}

TAGA is trained using an asymmetric actor--critic PPO objective augmented with auxiliary regularization losses. The overall optimization objective is

\begin{equation}
\mathcal{L}_{\mathrm{policy}}
=
\mathcal{L}_{\mathrm{PPO}}
+
c_v \mathcal{L}_{\mathrm{value}}
-
c_e \mathcal{H}(\pi_\theta)
+
\lambda_c \mathcal{L}_{\mathrm{con}}
+
\lambda_b \mathcal{L}_{\mathrm{roi}},
\end{equation}

where $\mathcal{L}_{\mathrm{PPO}}$ is the PPO surrogate loss, $\mathcal{L}_{\mathrm{value}}$ is the critic regression loss, $\mathcal{H}(\pi_\theta)$ is the policy entropy bonus, and $\mathcal{L}_{\mathrm{con}}$ and $\mathcal{L}_{\mathrm{roi}}$ denote the contrastive and gaze regularization losses, respectively.

\subsection{MoE--Terrain Contrastive Loss}

Following CMoE~\citep{ma2026cmoe}, we employ a contrastive objective to encourage terrain-aware soft routing in the MoE policy. Let $B$ denote the batch size and $\mathbb{K}$ denote the number of learnable prototypes in a shared prototype dictionary $C$. The contrastive pair is formed by the actor-gate embedding $e_i^{g}$ and the terrain embedding $e_i^{h}$ encoded from the full height scan. Both embeddings are normalized and compared through the prototype dictionary.

For an embedding $e$, we define its prototype prediction distribution and balanced assignment as

\begin{equation}
P(e)
=
\mathrm{softmax}(eC^\top/T_{\mathrm{con}}),
\qquad
Q(e)
=
\mathrm{Sinkhorn}(eC^\top),
\end{equation}

where $T_{\mathrm{con}}$ is a temperature parameter. The contrastive loss is

\begin{equation}
\mathcal{L}_{\mathrm{con}}
=
-\frac{1}{2B\mathbb{K}}
\sum_{i=1}^{B}
\left[
Q(e_i^{g})^\top \log P(e_i^{h})
+
Q(e_i^{h})^\top \log P(e_i^{g})
\right].
\end{equation}

This objective encourages the gating network to produce terrain-aware soft routing weights by aligning the gate embedding $e_i^{g}$ with the terrain embedding $e_i^{h}$ in the shared prototype space. This helps reduce terrain-invariant or nearly uniform expert mixtures and promotes functional differentiation among experts.

\subsection{Gaze Boundary Loss}

To prevent the learned gaze from degenerately selecting crops near the height-scan boundary, we penalize ROI centers that are too close to the edge of the normalized scan domain. For notational convenience, let
\begin{equation}
r_i=(r_i^x,r_i^y)\in[0,1]^2
\end{equation}
denote the normalized gaze location predicted by TAGA for the $i$-th sample in a mini-batch. Here, $r_i$ is a batch-level notation for the corresponding time-indexed prediction $r_t$ used in the policy rollout. 
Given a boundary margin  $m=0.05$, the boundary loss is defined as

\begin{equation}
\mathcal{L}_{\mathrm{roi}}
=
\frac{1}{2B}
\sum_{i=1}^{B}
\left(
[m-r_i^x]_{+} + [r_i^x-(1-m)]_{+}
+
[m-r_i^y]_{+} + [r_i^y-(1-m)]_{+}
\right),
\end{equation}

where $[\xi]_+=\max(\xi,0)$. This regularizer keeps the selected crop inside the valid height-scan region and prevents unstable edge fixation.

\paragraph{Symmetry Augmentation.}
We exploit the humanoid's left-right morphological symmetry by mirroring proprioceptive observations, height scans, AMP states, and actions. For depth observations, a horizontally flipped virtual camera view is used. Both original and mirrored samples are included in PPO updates, encouraging symmetric locomotion behaviors and improving training efficiency.

%===============================================================================

\clearpage

%===============================================================================

\end{document}